\documentclass[runningheads]{llncs}
\usepackage{graphicx}
\usepackage{comment}
\usepackage{amsmath,amssymb} 
\usepackage{color}

\usepackage{multirow}

\usepackage{subfigure}

\usepackage[marginal]{footmisc}

\newcommand{\Tref}[1]{Tab.~\ref{#1}}
\newcommand{\Eref}[1]{Eq.~(\ref{#1})}
\newcommand{\Fref}[1]{Fig.~\ref{#1}}

\begin{document}
	\pagestyle{headings}
	\mainmatter
	
	\title{Rethinking Spatially-Adaptive Normalization} 

	\titlerunning{ }
	\author{Zhentao Tan$^{1,}$\thanks{Equal contribution}, Dongdong Chen$^{2,*}$, Qi Chu$^{1}$, Menglei Chai$^{3}$, Jing Liao$^{4}$, Mingming He$^{5}$, Lu Yuan$^{2}$, Nenghai Yu$^{1}$}
	\authorrunning{ }
	%
	\institute{
		$^{1}$University of Science and Technology of China \quad 
		$^{2}$Microsoft Cloud AI \\
		$^{3}$Snap Inc.\quad$^4$City University of Hong Kong\quad $^5$University of Southern California\\}

\maketitle
\vspace{-2em}
\begin{figure}
	\begin{center}
		\includegraphics[width=0.99\linewidth]{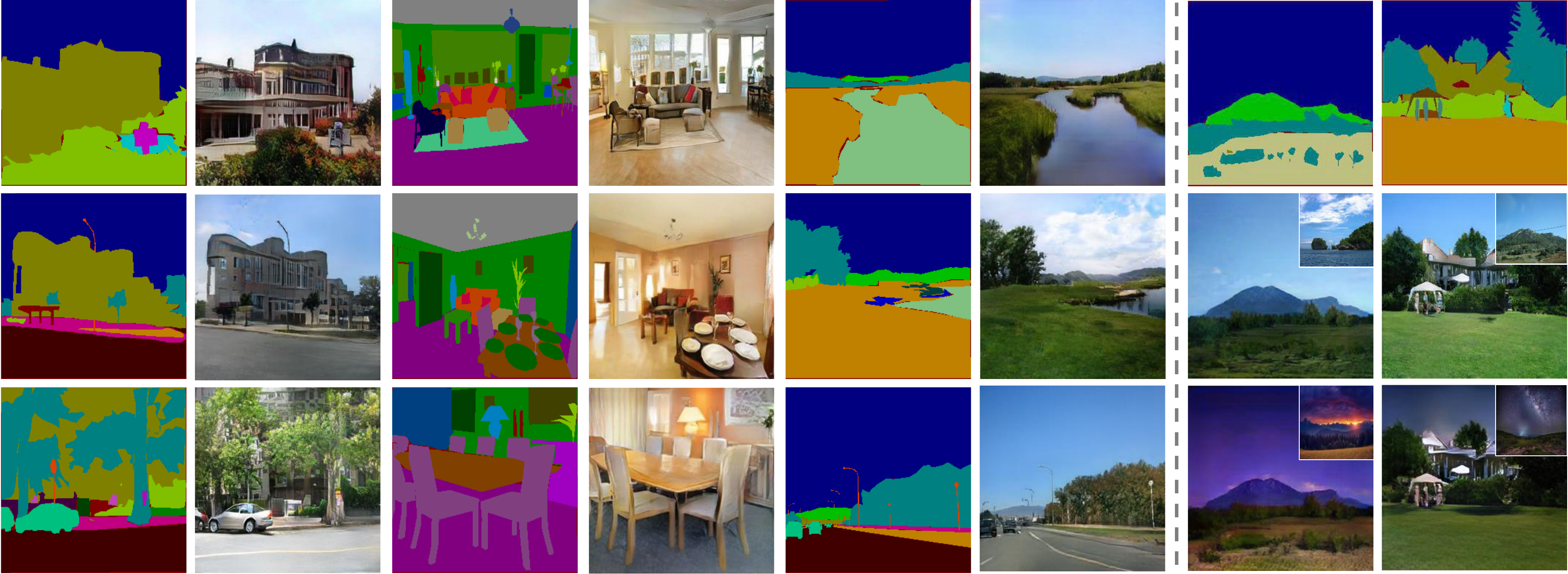}
		\caption{\footnotesize{Semantic synthesis results produced by our method. Our method can not only handle the synthesis from a pure semantic segmentation mask (left six columns) but also support controllable synthesis via different reference style images (right two columns). }}
		\label{fig:teaser}
	\end{center}
	\vspace{-4.5em}
\end{figure}

\begin{abstract}
	Spatially-adaptive normalization~\cite{park2019semantic} is remarkably successful recently in conditional semantic image synthesis, which modulates the normalized activation with spatially-varying transformations learned from semantic layouts, to preserve the semantic information from being washed away. Despite its impressive performance, a more thorough understanding of the true advantages inside the box is still highly demanded, to help reduce the significant computation and parameter overheads introduced by these new structures. In this paper, from a return-on-investment point of view, we present a deep analysis of the effectiveness of SPADE and observe that its advantages actually come mainly from its semantic-awareness rather than the spatial-adaptiveness. Inspired by this point, we propose class-adaptive normalization (CLADE), a lightweight variant that is not adaptive to spatial positions or layouts. Benefited from this design, CLADE greatly reduces the computation cost while still being able to preserve the semantic information during the generation. Extensive experiments on multiple challenging datasets demonstrate that while the resulting fidelity is on par with SPADE, its overhead is much cheaper than SPADE. Take the generator for ADE20k dataset as an example, the extra parameter and computation cost introduced by CLADE are only $4.57\%$ and $0.07\%$ while that of SPADE are $39.21\%$ and $234.73\%$ respectively. 
\end{abstract}

\section{Introduction}
Image synthesis has seen tremendous progress recently thanks to the advances in deep generative models. Latest successes, such as StyleGAN~\cite{karras2019style,karras2019analyzing}, are already capable of producing highly realistic images from random latent codes. Yet conditional image synthesis, the task of generating photo-realistic images conditioned on some input data, is still very challenging. In this work, we focus on semantic image synthesis, a specific conditional image generation task that aims at converting a semantic segmentation mask into a photo-realistic image.

To tackle this problem, some previous methods~\cite{isola2017image,wang2018high} directly feed the semantic segmentation mask as input to the conventional deep network architectures built by stacking convolution, normalization, and nonlinearity layers. However, as pointed out in~\cite{park2019semantic}, common normalization layers like instance normalization~\cite{ulyanov2016instance} tend to wash away the semantic information, especially for flat segmentation masks. To compensate for the information loss, a new spatially-adaptive normalization layer, SPADE~\cite{park2019semantic}, is proposed. It modulates the normalized activations through a spatially-adaptive transformation, which is learned from the input segmentation mask. Therefore, by replacing all the common normalization layers with SPADE blocks, the input semantic information can be successfully propagated throughout the network and achieve much better performance in terms of visual fidelity and spatial layout alignment.    

Despite its effectiveness, the true magic behind the success of SPADE has not been fully uncovered yet. Is \textit{spatial-adaptiveness} the sole or dominating reason for its superior performance, as suggested by the name of SPADE? Does there exist any better design that can improve efficiency without hurting the result quality? In this paper, we try to answer these questions with a deep analysis of SPADE. Our key observation is that \textit{semantic-awareness} may actually contribute much more than the \textit{spatial-adaptiveness}. In fact, since the two-layer modulation network used to regress the transformation parameters is so shallow, the resulting denormalization parameters are actually class-wise sparse across the entire semantic map, especially for high-resolution image synthesis tasks. Meanwhile, given that a SPADE block is placed after almost every convolutional layer, such redundancy is recurring multiple times in the generation pass, which can easily cause a heavy amount of unnecessary computation and parameter overhead.

Motivated by this observation, we propose a novel normalization layer, namely CLass-Adaptive (DE)normalization (CLADE). Different from the spatially adaptive solution of SPADE, CLADE instead uses the input semantic mask to modulate the normalized activations in a \emph{class-adaptive} manner. Specifically, CLADE is only adaptive to different semantic classes to maintain the crucial semantic-awareness property but independent of the spatial position, semantic shape or layout of the semantic mask. Benefited from this lightweight design, the implementation of CLADE is surprisingly simple and requires no extra modulation network. Therefore, its computation and parameter overhead are almost negligible compared with SPADE, making CLADE a better alternative to those conventional normalization layers.

To demonstrate the effectiveness and efficiency of CLADE, we conduct extensive experiments on multiple challenging datasets, including Cityscapes~\cite{cordts2016cityscapes}, COCO-Stuff~\cite{caesar2018coco} and ADE20k~\cite{zhou2017scene}. Without bells and whistles, just by replacing all the SPADE layers with CLADE, very comparable performance can be achieved with much smaller model size and much less computation cost. Take the generator for the ADE20k dataset as an example, the extra parameter and computation overhead of CLADE are only $4.57\%$ and $0.07\%$ while that of SPADE are $39.21\%$ and $234.73\%$. Like~\cite{park2019semantic}, our method also supports multi-modal and style-guided image synthesis, which enables controllable and diverse outputs. Some visual results are given in \Fref{fig:teaser}.

\section{Related Works}

\subsection{Conditional Image Synthesis}
Conditional image synthesis refers to the task of generating photo-realistic images conditioned on different types of input, such as 
texts~\cite{hong2018inferring,reed2016generative,xu2018attngan,zhang2017stackgan} and images~\cite{huang2018multimodal,isola2017image,liu2017unsupervised,zhu2017unpaired,oza2019semi,park2019semantic}. In this paper, we focus on a special form of conditional image synthesis that aims at generating photo-realistic images conditioned on input segmentation masks, called semantic image synthesis. 
For this task, many impressive works have been proposed in the past several years. One of the most representative work is Pix2Pix~\cite{isola2017image}, which proposes a unified image-to-image translation framework based on the conditional generative adversarial network. To further improve its quality or enable more functionality, many following works emerged, such as Pix2pixHD~\cite{wang2018high}, SIMS~\cite{qi2018semi}, and SPADE~\cite{park2019semantic}. With a thorough analysis of SPADE, a new efficient and effective normalization layer is introduced and able to achieve superior performance. 

\subsection{Normalization Layers}
In the deep learning era, normalization layers play a crucial role in achieving better convergence and performance, especially for deep networks. They follow a similar operating logic, which first normalizes the input feature into zero mean and unit deviation, then modulates the normalized feature with learnable modulation scale/shift parameters. Existing normalization layers can be generally divided into two different types: unconditional and conditional. Compared to unconditional normalization, the behavior of conditional normalization is not static and depends on the external input. Typical unconditional normalization layers include Local Response Normalization (LRN)~\cite{krizhevsky2012imagenet}, Batch Normalization (BN)~\cite{ioffe2015batch}, Instance Normalization (IN)~\cite{ulyanov2016instance}, Layer Normalization (LN)~\cite{ba2016layer} and Group Normalization (GN)~\cite{wu2018group}.

Conditional Instance Normalization (Conditional IN)~\cite{dumoulin2016learned} and Adaptive Instance Normalization (AdaIN)~\cite{huang2017arbitrary} are two popular conditional normalization layers originally designed for style transfer. To transfer the style from one image to another, they model the style information into the modulation scale/shift parameters. For semantic image synthesis, most works before~\cite{park2019semantic} just leveraged unconditional normalization layers BN or IN in their networks. Recently, Park \emph{et al.}~\cite{park2019semantic} points out that common normalization layers used in existing methods tend to "wash away" semantic information when applied to flat segmentation masks. To compensate for the lost information, they innovatively propose a new spatially-adaptive normalization layer named SPADE. Different from common normalization layers, SPADE adds the semantic information back by making the modulation parameters be the function of semantic mask in a spatially-adaptive way. Experiments also demonstrate SPADE can achieve state-of-the-art performance in semantic image synthesis. In this paper, we rethink the properties of SPADE via deep analysis and find that the semantic-awareness is the essential property leading to the superior performance of SPADE rather than the spatially-adaptiveness. Moreover, the computation and parameter complexity of SPADE is also very high. Based on this observation, a new normalization layer CLADE is further proposed. It can achieve comparable performance as SPADE but with very negligible cost.

\section{Semantic Image Synthesis}
Conditioned on a semantic segmentation map $m\in\mathbb{L}^{H\times W}$, semantic image synthesis aims to generate a corresponding high-quality photorealistic image $I$. Here, $\mathbb{L}$ is the set of class integers that denote different semantic categories. $H$ and $W$ are the target image height and width.

Most vanilla synthesis networks, like Pix2Pix~\cite{isola2017image} and Pix2PixHD~\cite{wang2018high}, adopt a similar network structure concatenating repeated blocks of convolutional, normalization and non-linear activation layers. Among them, normalization layers play a crucial role for better convergence and performance. They can be generally formulated as:
\begin{equation}
\begin{split}
\hat{x}^{in}_{i,j,k} &= \frac{x^{in}_{i,j,k}-\mu_{i,j,k}}{\sigma_{i,j,k}},\\
x^{out}_{i,j,k} &= \gamma_{i,j,k} \hat{x}^{in}_{i,j,k} + \beta_{i,j,k},
\end{split} 
\end{equation}
with the indices of (width, height, channel) denoted as $i,j,k$. In what follows, for the simplicity of notation, these subscripts will be omitted if the variable is independent of them. Specifically, the input feature $x^{in}$ is first normalized with the mean $\mu$ and standard deviation $\sigma$ (normalization step), and then modulated with the learned scale $\gamma$ and shift $\beta$ (modulation step). For most common normalization layers such as BN~\cite{isola2017image} and IN~\cite{ulyanov2016instance}, all four parameters are calculated in a channel-wise manner (independent of $i,j$), with the modulating parameters $\gamma$ and $\beta$ independent of $x^{in}$.

\subsection{Revisit Spatially-Adaptive Normalization}
One common issue of the aforementioned normalization layers, as pointed out in~\cite{park2019semantic}, is that they tend to wash away the semantic information on sparse segmentation masks $m$ during image synthesis. Take IN as an example, if the network input is a piece-wise constant mask $m$, the corresponding convolution output $x^{in}_k$ will have a similar spatial sparsity. Though $x^{in}_k$ may vary for different labels, the normalized $\hat{x}^{in}_k$ will become all zeros after subtracting the channel-wise mean $\mu_k$ ($\mu_k=x^{in}_k$). Since $\gamma, \beta$ are also independent of $x^{in}$ and $m$, most of the semantic information will get easily smeared.

\begin{figure}[tp]
	\centering
	\includegraphics[width=0.95\linewidth]{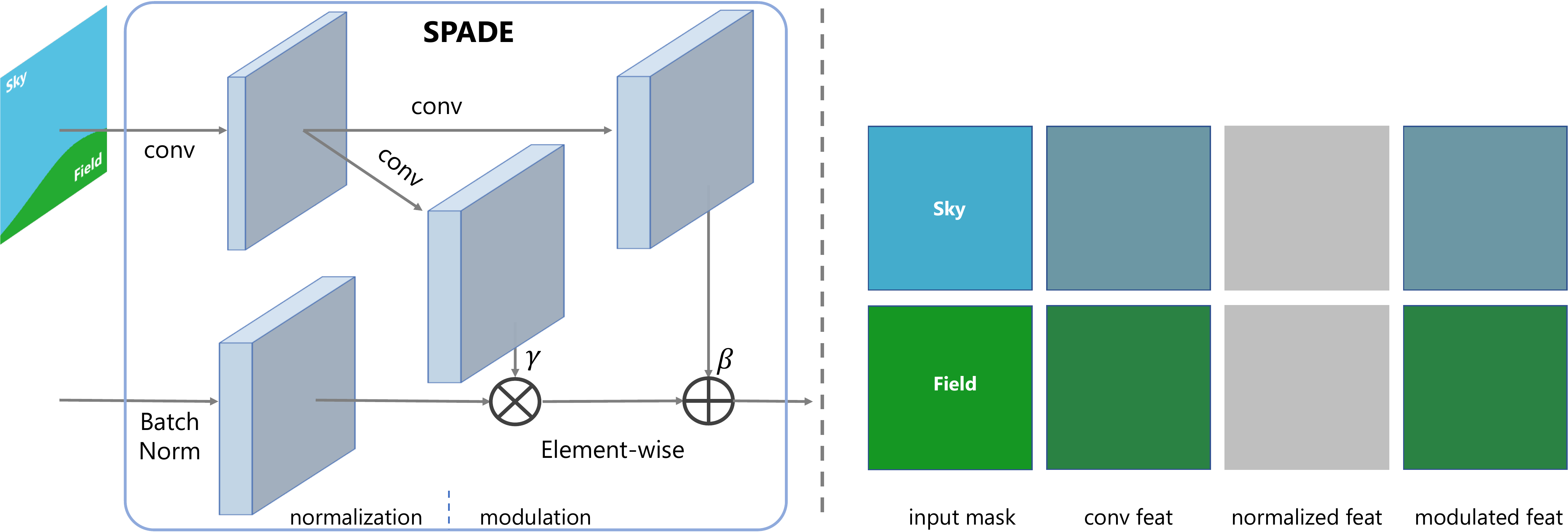}
	\caption{Working principle of SPADE (left) and one example (right) to show why SPADE can preserve semantic information from being washed away. By modeling the modulation parameters $\gamma,\beta$ as the function of input semantic mask, it can add the semantic information lost in the normalization step back.}
	\label{fig:SPADE}
\end{figure}

Motivated by this observation, a new spatially-adaptive normalization layer, namely SPADE, is proposed in~\cite{park2019semantic}. By making the modulation parameters $\gamma$ and $\beta$ functions of the input mask $m$, i.e., $\gamma_{i,j,k}(m)$ and $\beta_{i,j,k}(m)$, the semantic information, which is lost after the normalization step, will be added back during the modulation step. The functions of $\gamma_{i,j,k}(\cdot)$ and $\beta_{i,j,k}(\cdot)$ are both implemented with a shallow modulation network consisting of two convolutional layers, as illustrated in \Fref{fig:SPADE}. By replacing all the normalization layers with SPADE, the generation network proposed in~\cite{park2019semantic} can achieve much better synthesis results than previous state-of-the-art methods like Pix2PixHD~\cite{wang2018high}. 

\subsubsection{Working principle analysis.} The advantages of SPADE come mainly from two important properties: \emph{spatial-adaptiveness} and \emph{semantic-awareness}. The former means the modulation parameters $(\gamma,\beta)$ are spatially varying in a pixel-wise manner, while the latter property means that $(\gamma,\beta)$ are dependent on semantic classes to bring back the lost information. It may appear that the spatial-adaptiveness is more important, which is also suggested by the name of SPADE. However, through our analysis below, we argue that the semantic-awareness is the de facto main contributor to SPADE. 

\begin{figure}[tp]
	\centering
	\includegraphics[width=0.75\linewidth]{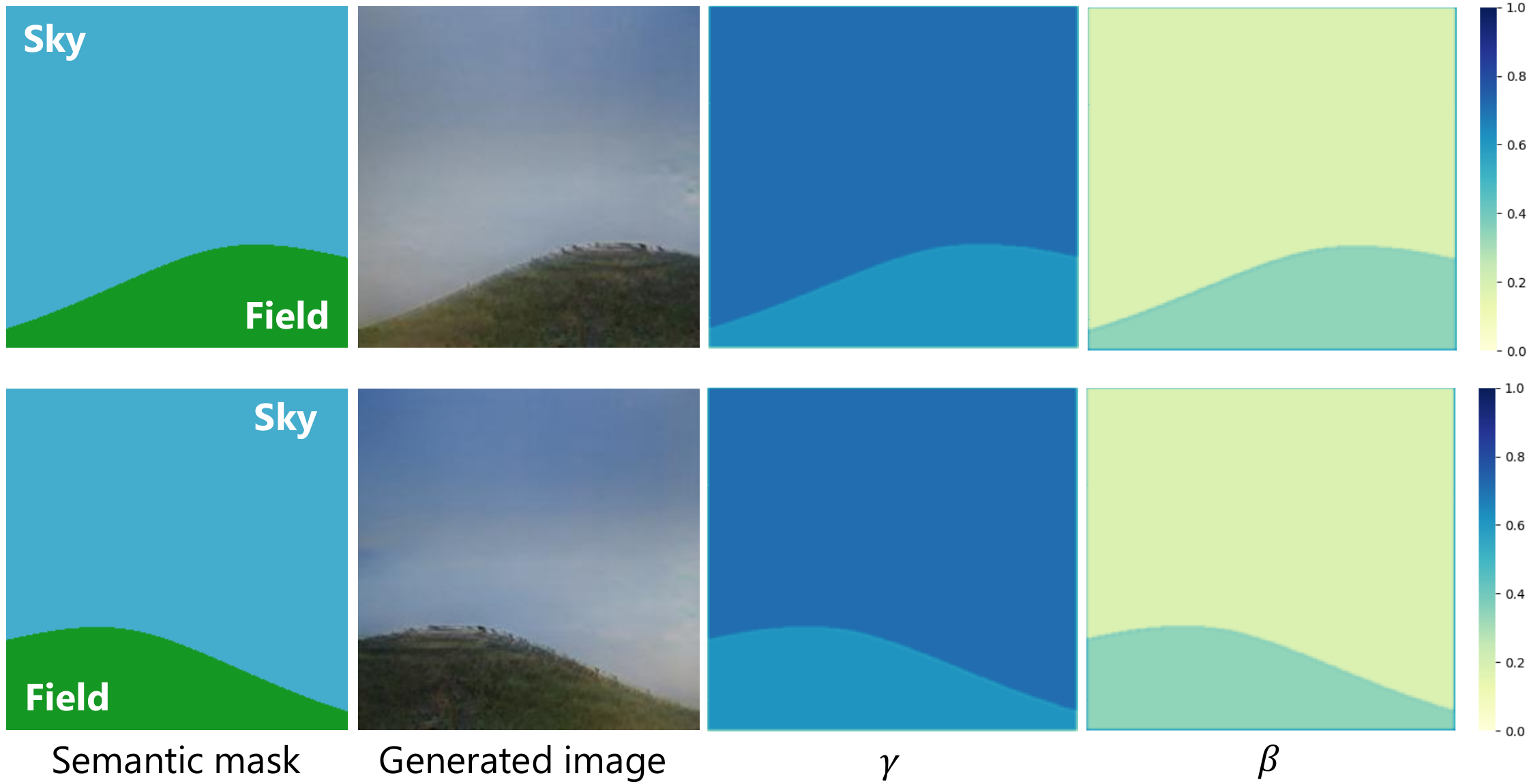}
	\caption{Visualization of learned modulation parameters $\gamma,\beta$ for two example semantic masks from the ADE20k dataset, where the original pretrained SPADE generator is used. Obviously, $\gamma,\beta$ for the same semantic class are almost identical within each semantic region or across different masks.}
	\label{fig:feature}
\end{figure}

First, let us reuse the piece-wise constant mask $m$ as an example. Since the modulation network in SPADE does not include any normalization layer, $\gamma(m),\beta(m)$ vary across different semantic classes, but are still uniform within a same class. In other words, $\gamma(m),\beta(m)$ are semantic-aware but not spatially-adaptive in fact. In \Fref{fig:feature}, we further show two examples with the masks $m$ from the ADE20k validation dataset~\cite{zhou2017scene}, which both consist of two semantic labels ``Sky" and ``Field" but with different shapes. We visualize the intermediate parameters of $\gamma$ and $\beta$ with the original pre-trained SPADE generator. It can be easily observed that $\gamma,\beta$ are almost identical within each semantic region, except for the boundary area which is negligible due to the shallowness of the modulation network. In fact, for any two regions sharing the same semantic class within one input mask or even across different input masks, their learned $\gamma, \beta$ will also be almost identical. This phenomenon is more common for high-resolution image synthesis because the sizes of most regions are more likely to be larger than the receptive field of the two-layer modulation network. This further confirms that the semantic-awareness may be the underlying key reason for the superior performance of SPADE compared with the spatially-adaptiveness.  

\subsection{Class-Adaptive Normalization}
\begin{figure}[tp]
	\centering
	\includegraphics[width=0.95\linewidth]{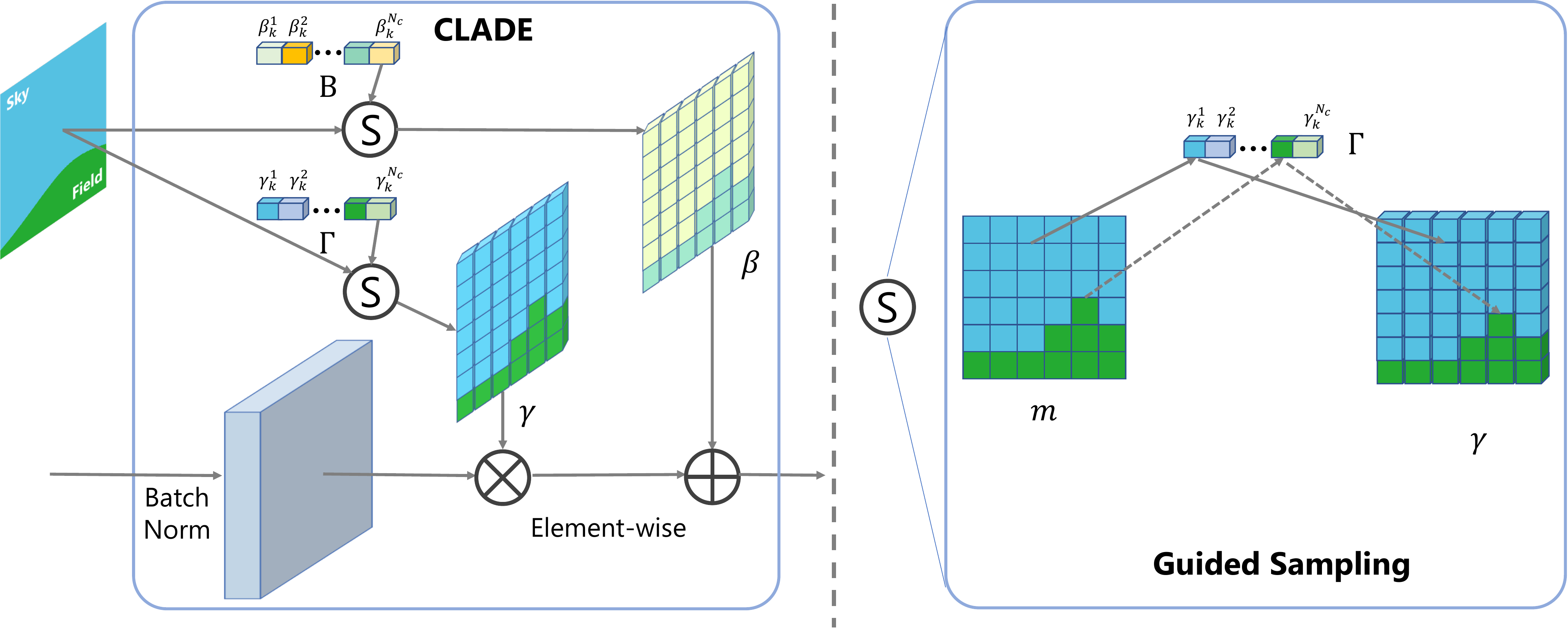}
	\caption{The illustration diagram of our class-adaptive normalization layer CLADE. Unlike SPADE, it does not introduce any external modulation network but only uses a cheap guided sampling operation to sample class-adaptive modulation parameters for each semantic region. }
	\label{fig:CLADE}
\end{figure}

Inspired by the above observation, we propose a novel conditional normalization layer, called CLass-Adaptive (DE)normalization (CLADE), as shown in \Fref{fig:CLADE}. Inheriting the idea of semantic information compensation from SPADE, the modulation parameters $(\gamma,\beta)$ in CLADE are also adaptive to the semantic input of $m$. However, instead of adopting the pixel-wise spatial-adaptiveness as in SPADE, CLADE is spatially-invariant and only adaptive to different semantic classes. More specifically, $(\gamma, \beta)$ in CLADE vary on the corresponding semantic classes to maintain the essential property of semantic-awareness, but are independent of any spatial information including the position, semantic shape, or layout of $m$.

Therefore, rather than learning modulation parameters through an extra modulation network like SPADE, we directly maintain a modulation parameter bank for CLADE and optimize it as regular network parameters. Assuming the total class number in $\mathbb{L}$ to be $N_c$, the parameter bank consists of $N_c$ channel-wise modulation scale parameters $\Gamma=(\gamma^1_k,...,\gamma^{N_c}_k)$ and shift parameters $B=(\beta^1_k,...,\beta^{N_c}_k)$. During training, given an input mask $m$, we fill each semantic region of class $l$ with its corresponding modulation parameter $\gamma_k^l, \beta_k^l$ to generate dense modulation parameter tensors $\overrightarrow{\gamma}$ and $\overrightarrow{\beta}$ respectively. We call this process \emph{Guided sampling} in \Fref{fig:CLADE}. 

In fact, CLADE can also be regarded as a generalized formulation of some existing normalization layers. If $\gamma_k^{l_1}\equiv\gamma_k^{l_2}$ and  $\beta_k^{l_1}\equiv\beta_k^{l_2}$ for any $l_1,l_2\in\mathbb{L}$, CLADE becomes BN~\cite{ioffe2015batch}. And if we make the modulation tensor $\overrightarrow{\gamma}$ and $\overrightarrow{\beta}$ both spatially uniform, and replace the BN mean and std statistics with the corresponding ones from IN, we arrive at Conditional IN.

\subsection{Computation and Parameter Complexity Analysis} 

\subsubsection{Analysis for SPADE.} 
\begin{figure}[tp]
	\centering
	\includegraphics[width=\linewidth]{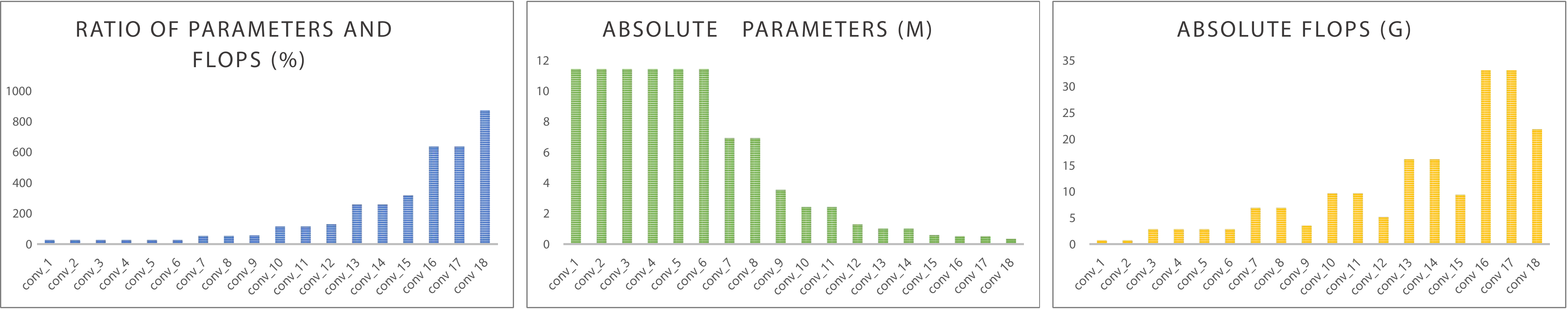}
	\caption{Left: The relative ratios of the parameter and FLOPs between each SPADE and its precedent convolutional layer in the original SPADE generator. Middle and right: absolute parameter number and FLOPs for different layers. $x-$axis represents layer index from shallow to deep.}
	\label{fig:ratio}
\end{figure}

In the original SPADE generator backbone~\cite{park2019semantic}, a SPADE block is placed after almost every convolution to replace the conventional normalization layer. For the sake of convenience, we denote the input and output channel number of the precedent convolutional layer as $C_{in}, C_{out}$ and its kernel size as $k_c$. For its modulation network, we simply assume a same kernel size $k_m$ and intermediate channel number $C_m$ are used for all convolutional layers. Therefore, the parameter numbers for the precedent convolutional layer $P_{conv}$ and the SPADE block $P_{spade}$ are calculated as:
\begin{equation}
P_{conv} = k_c^2*C_{in}*C_{out}, \quad P_{spade} = k_m^2*(N_c*C_m + 2*C_m*C_{out}).
\end{equation}
With the default implementation settings of SPADE, we have $k_c=k_m=3$, so the parameter ratio between both networks is:
\begin{equation}
P_{spade}/P_{conv}=\frac{N_c*C_m + 2*C_m*C_{out}}{C_{in}*C_{out}}.
\end{equation}
This to say, the extra parameter introduced by SPADE becomes a significant overhead, especially when $N_c,C_m$ are relatively large ($C_m=128$ by default in SPADE). Take the ADE20k dataset~\cite{zhou2017scene} as an example, which contains 151 classes ($N_c=151$). On image resolution of $256\times 256$, the SPADE generator consists of 7 SPADE residual blocks. We show the parameter ratio $P_{spade}/P_{conv}$ of each convolutional layer In \Fref{fig:ratio}. It can be seen that SPADE indeed brings significant parameter overhead to all the convolutional layers. This becomes even severer when the network goes deeper, since $C_{in}$ is designed to be smaller for larger feature resolution. The ratios for some layers can even surpass $600\%$. Taking all the convolutional layers in SPADE generators into consideration, the average ratio is about $39.21\%$.

Besides the parameter numbers, we also analyze the computation complexity. Here, we use the popular floating-point operation per second (FLOPs) as the metric. Since the convolutional layers within the modulation network dominate the computation cost of SPADE, the FLOPs of both the SPADE block $F_{spade}$ and the precedent convolutional layer $F_{conv}$ can be simply approximated as:
\begin{equation}
F_{conv} = k_c^2*C_{in}*C_{out}*H*W, \quad F_{spade} = k_m^2*(N_c*C_m + 2*C_m*C_{out})*H*W,
\end{equation}
where $H,W$ are the width and height of the output feature respectively. Therefore, the FLOPs ratio $F_{spade}/P_{conv}$ is identical to the parameter ratio shown in \Fref{fig:ratio}. However, different from the parameter number, with the increasing feature resolutions, the absolute FLOPs are relatively larger in deeper layers, which makes the computation overhead even worse. Using the same ADE20k example, the average extra FLOPs ratio introduced by SPADE is about $234.73\%$, which means the computation cost of SPADE is even heavier than the precedent convolutional layers. More importantly, it is now popular to adopt very large synthesis networks to ensure good performance, which is already consuming a surprisingly large amount of parameter spaces and computation resources, and SPADE will significantly worsen this situation, which might be unaffordable in many cases.    

\subsubsection{Analysis for CLADE.} Compared to SPADE, our CLADE does not require any extra modulation network to regress the modulation parameters. Specifically, the corresponding numbers of parameter and FLOPs of CLADE are:
\begin{equation}
P_{clade} = 2N_c*C_{out}, \quad F_{clade} = C_{out}*H*W.
\end{equation}
Here we regard the value assignment operation as one float-point operation. Similar to SPADE, if every convolutional layer is followed by one CLADE layer, the relative parameter and FLOPs ratio will be:
\begin{equation}
P_{clade}/P_{conv}=\frac{2*N_c}{k_c^2*C_{in}}, \quad F_{clade}/F_{conv}=\frac{C_{out}}{k_m^2*C_{in}*C_{out}}.
\end{equation}
In most existing synthesis networks, the above ratios are extremely small. For example, with the same backbone as the above SPADE generator for the ADE20k dataset, the average ratios for parameter and FLOPs are only $4.57\%$ and $0.07\%$, respectively. Therefore, compared to SPADE, the parameter and computation overhead of CLADE are both negligible, which is friendly to practical scenarios regarding both training and inference. Despite its simplicity and efficiency, we demonstrate that it can still achieve comparable performance as SPADE with extensive experiments in Sec.~\ref{sec:exp}. 

\subsection{CLADE Generator}
The CLADE generator follows the similar network architecture of the SPADE generator, but with all the SPADE layers are replaced by CLADE. Basically, it adopts several residual blocks with upsampling layers and progressively increases the output resolution. And since the semantic information from $m$ is already encoded by CLADE, there is no encoder part in the generator either. For multi-modal synthesis, we follow the strategy as~\cite{park2019semantic} which attaches an extra encoder that encodes an image into a random vector. This vector extracts the style information from the image and is then fed into the CLADE generator as the style guidance. For more details, we encourage the readers to review SPADE~\cite{park2019semantic}.

\section{Experiments}
\label{sec:exp}
\subsection{Implementation details}
We follow the same training setting as SPADE~\cite{park2019semantic}. In detail, the generator is trained with the same multi-scale discriminator and loss function used in SPADE~\cite{park2019semantic}. By default, the Adam~\cite{kingma2015adam} ($\beta_1=0,\beta_2=0.9$) is used with the total epoch number of 200. The learning rates for the generator and discriminator are set to 0.0001 and 0.0004, respectively. We evaluate the model every 10 epochs and select the model with the best performance. All of the experiments and analyses are conducted on TITAN XP and V100 GPUs with the Pytorch framework~\cite{paszke2019pytorch}.

\subsection{Datasets and metrics}

\subsubsection{Datasets.} Experiments are conducted on four popular segmentation datasets.
\begin{itemize}
	\item $ADE20k$ dataset~\cite{zhou2017scene} consists of 25,210 images (20,210 for training, 2,000 for validation and 3,000 for testing). The images in $ADE20k$ dataset cover a wide range of scenes and object categories, including a total of 150 object and stuff classes. 
	\item $ADE20k$-$outdoor$ dataset is a subset of the ADE20K dataset that only contains outdoor scenes, which is used in several previous works~\cite{park2019semantic,qi2018semi}. Unlike previous work, we directly select the images containing categories such as sky, trees, and sea without manual inspection. There are 9,649 training images and 943 validation images.
	\item COCO-Stuff~\cite{caesar2018coco} has the same number of images as COCO dataset~\cite{lin2014microsoft}, but augments COCO by adding dense pixel-wise stuff annotations. It has 118,000 training images and 5,000 validation images with 182 semantic classes.
	\item Cityscapes dataset~\cite{cordts2016cityscapes} is a widely used dataset for semantic image synthesis. The images are taken from street scenes of German cities. It contains 2,975 high-resolution training images and 500 validation images. The number of annotated semantic classes is 35. Recently, some works~\cite{wang2018video,qi2018semi,wang2018high} have been able to generate realistic semantic images on this dataset.
\end{itemize}

\subsubsection{Metrics.}We leverage the protocol from previous works~\cite{chen2017photographic,wang2018high} for evaluation, which is also used in the SPADE baseline~\cite{park2019semantic}. In details, we will run semantic segmentation algorithms on the synthesized images, and evaluate the quality of the predicted semantic mask. The intuition behind this metric is that, if the synthesized images are realistic enough, the semantic segmentation mask predicted by existing semantic segmentation models should be close to the ground truth. To measure the segmentation accuracy, two popular metrics, mean Intersection-over-Union (mIoU) and pixel accuracy (accu) metrics are used in the following experiments. And for different datasets, we selected corresponding state-of-the-art segmentation models: UperNet101~\cite{xiao2018unified,github-upernet} for $ADE20k$ and $ADE20k$-$outdoor$, DeepLabv2~\cite{chen2017deeplab,github-deeplab} for COCO-Stuff and DRN~\cite{yu2017dilated,github-drn} for Cityscapes. We also use Fr\'echet Inception Distance (FID)~\cite{heusel2017gans} to measure the distance of distributions between synthesized results and of real images. As a reference, we calculate these above metrics with real images on all datasets. 

\subsection{Quantitative and qualitative comparison}
\begin{table}[t]
	\centering
	\caption{Comparison with recent state-of-the-art models. For mIoU and pixel accuracy (accu), higher is better. For FID, lower is better.}
	\begin{tabular}{c|ccc|ccc|ccc|ccc}
		\hline
		& \multicolumn{3}{c|}{$ADE20k$} & \multicolumn{3}{c|}{$ADE20k$-$outdoor$} & \multicolumn{3}{c|}{COCO-Stuff} & \multicolumn{3}{c}{Cityscapes} \\
		\hline
		Method & mIoU & accu & FID & mIoU & accu & FID & mIoU & accu & FID & mIoU & accu & FID \\
		\hline
		Reference & 42.76 & 81.23 & 0.0 & 25.85 & 81.96 & 0.0 & 39.05 & 66.77 & 0.0 & 71.35 & 95.32 & 0.0\\
		\hline
		SIMS~\cite{qi2018semi} & N/A & N/A & N/A & N/A & N/A & N/A & N/A & N/A & N/A & 47.47 & 85.69 & 49.13\\
		Pix2pixHD~\cite{wang2018high} & 27.27 & 72.61 & 50.34 & 14.89 & 76.70 & 71.26 & 18.01 & 52.21 & 77.43 & 56.44 & 92.99 & 67.50 \\
		SPADE~\cite{park2019semantic} & 36.63 & 78.28 & 36.77 & 19.30 & 80.44 & 54.16 & 35.44 & 67.07 & 29.42 & 67.57 & 94.83 & 55.15\\
		CLADE & 35.43 & 77.36 & 38.56 & 18.71 & 80.77 & 55.81 & 35.19 & 65.84 & 30.55 & 65.16 & 94.67 & 55.81\\
	\end{tabular}
	\label{tab:performance} 
\end{table}{}
To demonstrate the effectiveness of our method, we not only compare our CLADE with the SPADE baseline ~\cite{park2019semantic} but also include two previous semantic image synthesis methods SIMS~\cite{qi2018semi} and Pix2pixHD~\cite{wang2018high}. SIMS is a semi-parametric synthesis method and needs to retrieve similar patches in an image patch library during runtime. Therefore, it is more computation-consuming. For a fair comparison, we directly adopt result images of SIMS provided by the authors in all following comparison experiments. For Pix2pixHD and SPADE, we use the codes and settings provided by the authors to train all the models.
The resolution of images is set to $256\times256$ except for Cityscapes, which is set to $1024\times512$. Since the number of training images on COCO-Stuff dataset is too large, we reduce the number of training epochs on this dataset to 50.

\vspace{1em}
\noindent \textbf{Quantitative results.} As shown in \Tref{tab:performance}, our method can achieve very comparable performance with the state-of-the-art baseline SPADE on all the datasets. For example, on the COCO-Stuff dataset, the proposed CLADE achieves a mIoU score of 35.19, only 0.25 less than SPADE. When compared to Pix2pixHD, CLADE improves it by more than $17\%$. As for the FID score, our CLADE is also very close to SPADE and much better than other methods. 

\begin{figure}[tp]
	\centering
	\includegraphics[width=122mm]{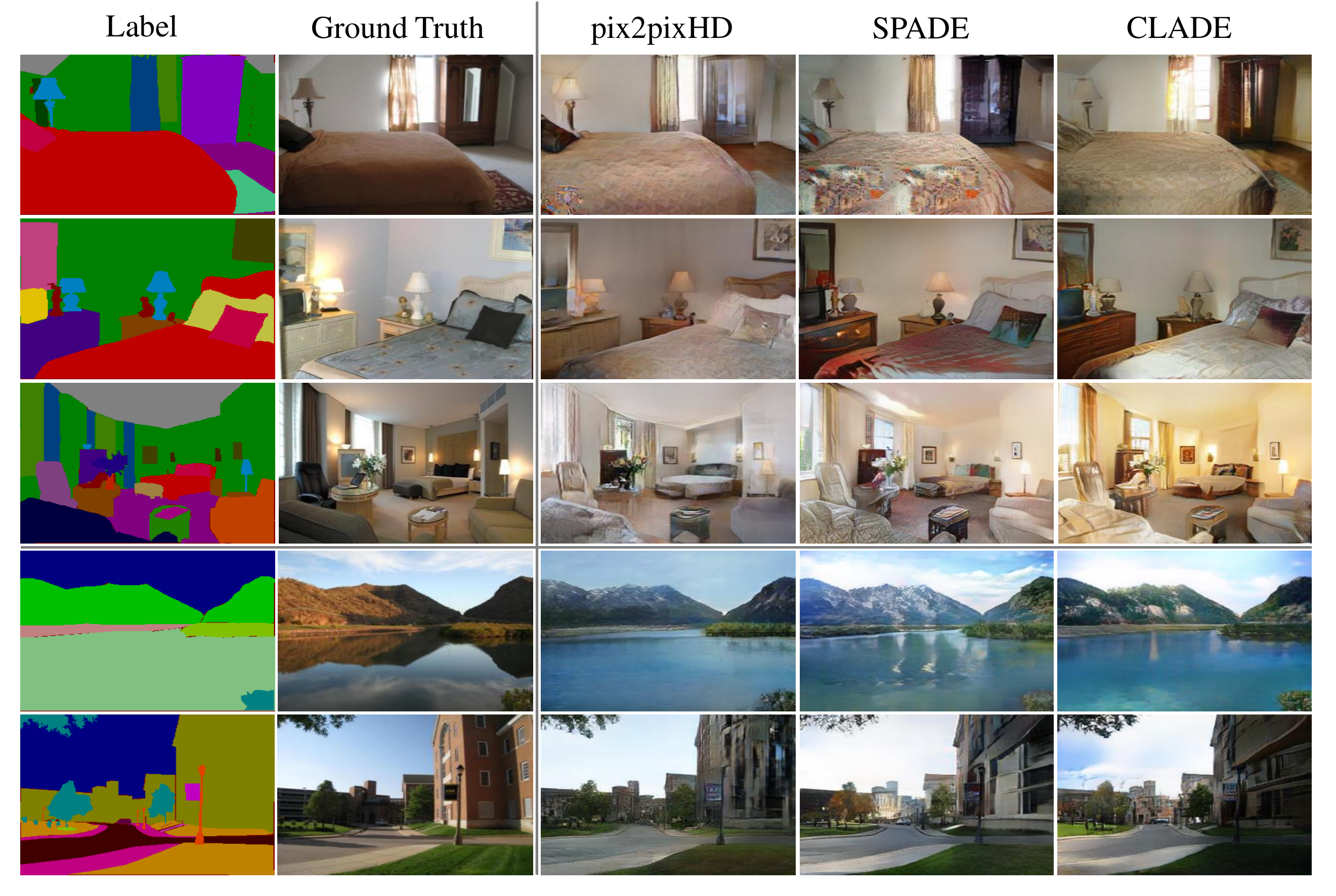}
	\caption{Visual comparison results on the $ADE20k$ (top three rows) and $ADE20k$-$outdoor$ (bottom two rows) dataset. It shows that images generated by our method are very comparable or even slightly better than SPADE. Compared to Pix2pixHD, SPADE and CLADE are overall more realistic.}
	\label{fig:results_ade}
\end{figure}

\begin{figure}[tp]
	\centering
	\includegraphics[width=122mm]{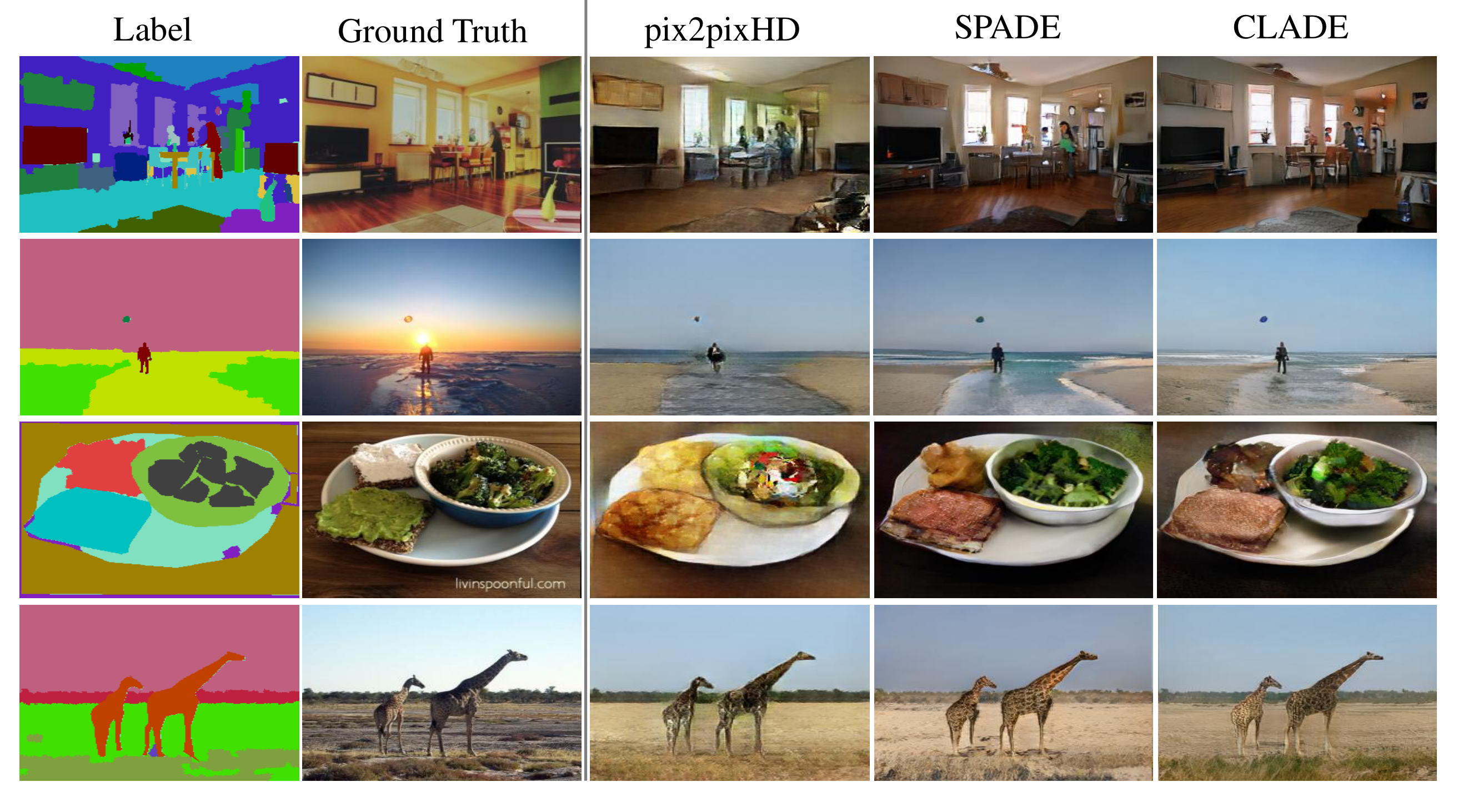}
	\caption{Visual comparison results on the challenging COCO-Stuff dataset. Though very diverse categories and small structures exist in this dataset, our method can still work very well and generate very high-fidelity results.}
	\label{fig:results_coco}
\end{figure}

\begin{figure}[tp]
	\centering
	\includegraphics[width=\linewidth]{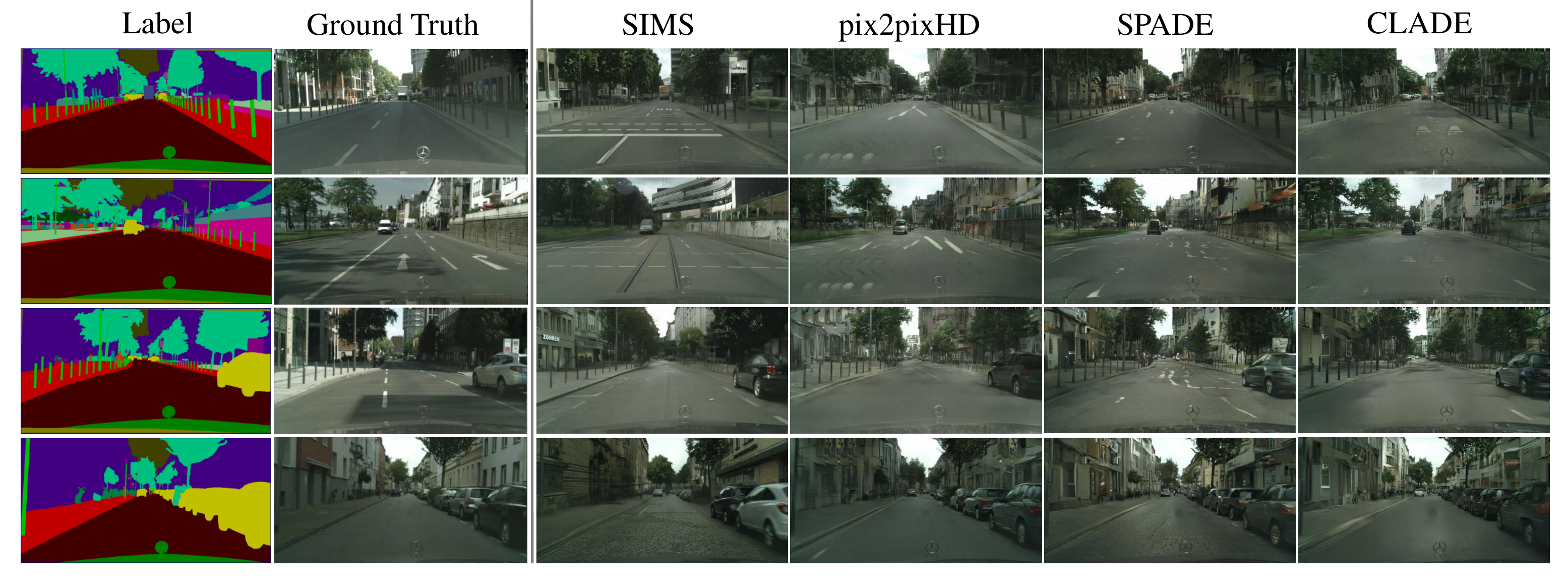}
	\caption{High-resolution synthesis ($1024\times 512$) results on the Cityscapes dataset. Our method produces realistic images with good spatial alignment and semantic meaning.}
	\label{fig:results_city}
\end{figure}

\vspace{1em}
\noindent \textbf{Qualitative results.} Besides the above quantitative comparison, we further provide some qualitative comparison results on the four different datasets. In detail, \Fref{fig:results_ade} shows some visual results on some indoor cases on the ADE20k dataset and outdoor cases on the ADE20k-outdoor dataset. Despite the simplicity of our method, it can generate very high-fidelity images that are comparable to the ones generated by SPADE. In some cases, we find our method is even slightly better than SPADE. In contrast, because of semantic information lost problem existing in common normalization layers, the results generated by Pix2pixHD are worse than both SPADE and our CLADE. In \Fref{fig:results_coco}, some visual results on the COCO-Stuff dataset are provided. Compared to ADE20k, COCO-stuff has more categories and contains more small objects, so it is more challenging. However, our method can still work very well and generate high-fidelity results. According to results in \Fref{fig:results_city}, a similar conclusion can also be drawn for higher-resolution semantic image synthesis on the Cityscapes dataset ($1024\times512$).

\vspace{1em}
\noindent \textbf{User study.} Considering judging the visual quality of one image is often subjective, we further conduct a user study to compare the results generated by different methods. Specifically, we give the users two synthesis images generated from the same semantic mask from two different methods (our method and baseline method). and ask them ``which is more realistic". To allow more careful comparisons, no time limit is applied. And for each pairwise comparison, we randomly choose 40 results for each method and involve 20 users. In \Tref{tab:userstudy}, we report the evaluation results on four different datasets. According to the results, we find that users have no obvious preference between our CLADE and SPADE, which once again demonstrates the comparable performance to SPADE. When compared to the results of Pix2pixHD, users clearly prefer our results on all the datasets, especially on the challenging COCO-Stuff dataset.

\begin{table}[t]
	\centering
	\caption{User study. The numbers indicate the percentage of users who favor the results of the proposed
		CLADE over the competing method.}
	\begin{tabular}{c|c|c|c|c}
		\hline
		Method & $ADE20k$ & $ADE20k$-$outdoor$ & COCO-Stuff & Cityscapes \\
		\hline
		CLADE vs. SPADE & 48.375 & 57.000 & 55.000 & 53.375 \\
		CLADE vs. Pix2pixHD & 68.375 & 73.375 & 95.000 & 57.500 \\
	\end{tabular}
	\label{tab:userstudy}
\end{table}

\begin{table}[t]
	\centering
	\caption{Complexity analysis. Params denotes the number of parameters, while FLOPs is the floating-point operations per second.}
	\begin{tabular}{c|ccc|ccc|ccc}
		\hline
		& \multicolumn{3}{c|}{ADE20k} & \multicolumn{3}{c|}{COCO-Stuff} & \multicolumn{3}{c}{Cityscapes} \\
		\hline
		\multirow{2}{*}{Method} & Params & FLOPs & Runtime& Params & FLOPs & Runtime & Params & FLOPs & Runtime\\
		& (M) & (G) & (s) & (M) & (G) & (s) & (M) & (G) & (s)\\ 
		\hline
		Pix2pixHD & 182.9 & 99.3 & 0.041 & 183.0 & 105.9 & 0.045 & 182.5 & 603.7 & 0.108\\
		SPADE & 96.5 & 181.3 & 0.042 & 97.5 & 191.0 & 0.044 & 93.0 & 1124.0 & 0.196 \\
		CLADE & 71.4 & 42.2 & 0.024 & 72.4 & 42.2 & 0.025 & 67.8 & 300.9 & 0.089\\
	\end{tabular}
	\label{tab:params}
\end{table}{}

\subsection{Comparisons on model size and computation cost.} Besides the theoretical complexity analysis in the above section, we provide the detailed model size and computation cost of the generator on different datasets in \Tref{tab:params}. It can be seen that the proposed CLADE generator significantly reduces the parameter number and computational complexity of the original SPADE generator on all the datasets. In detail, the parameter number in our CLADE generator is about $74\%$ of that in the original SPADE generator and $39\%$ of that in Pix2pixHD. As for the computation complexity in terms of FLOPs, CLADE generator is about $4\times$ cheaper than SPADE generator and $2\times$ cheaper than Pix2pixHD. Considering the GPU computation is often overqualified for single image processing, the real runtime speedup is less significant than FLOPs, but we still observe CLADE has a $2$-times speedup.

\begin{figure}
	\centering
	\includegraphics[width=0.98\linewidth]{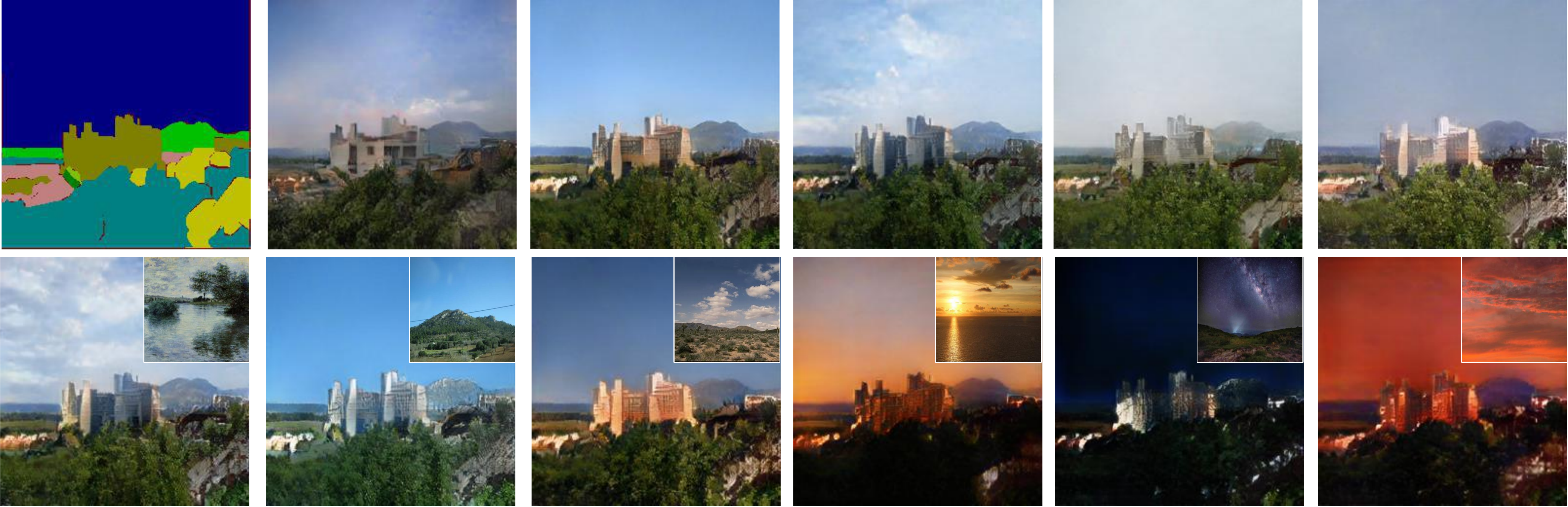}
	\caption{Multi-modal semantic synthesis results guided by different noise vectors (top row) or reference style images (bottom row). Obviously, our method can produce very diverse realistic images.}
	\label{fig:multi_modal}
\end{figure}
\subsection{Multi-modal and style-guided synthesis}
By feeding different noise vectors or reference style images, our method can produce diverse and controllable realistic images. As for the latter case, an extra style encoder needs to be added before the generator network. In \Fref{fig:multi_modal}, we show some multi-modal synthesis results by feeding different noise vectors in the top row and reference images as guidance in the bottom row.

\section{Conclusion}
In this paper, we rethink the popular spatially-adaptive normalization layer SPADE in semantic image synthesis.  By preserving the semantic information from being washed away, SPADE helps achieve much more impressive visual fidelity than previous normalization layers. However, the true advantages of SPADE are still unclear. Moreover, as the alternative of common normalization layers, SPADE will introduce very significant  computation and parameter overhead. Through deep analysis, we observe that the essential advantage of SPADE may come from semantic-awareness rather than spatially-adapativeness as originally suggested in their paper\cite{park2019semantic}. Motivated by this observation, we further design a more lightweight normalization layer, called ``class-adaptive normalization". Compared to SPADE, CLADE can achieve comparable results on different datasets, but its extra parameter and computation is almost negligible. Therefore, we believe it is a better alternative for common normalization layers. 
\bibliographystyle{splncs04}
\bibliography{egbib}

\clearpage
\appendix
\section{Using instance maps}
The proposed CLADE also supports the additional input of instance maps to better distinguish the different instances of the same categories. Similar to pix2pixHD and SPADE, we first calculate the edge map $E$ from the instance map (`edge' and `non-edge' are represented as `1' and `0'). Then, the edge map is modulated by:
\begin{equation}
\hat{E} = \gamma_{c} * E + \beta_{c}
\label{eq:edge}
\end{equation}
where $\hat{E}$ is the modulated edge map. $\gamma_c$ and $\beta_c$ are two constant float point numbers that can be learned as regular parameters. To embed edge information, we concatenate the modulated $\hat{E}$ with the feature maps $x^{out}$ modulated by the CLADE layer along the channel dimension, then feed them into the following layers. Since only two constant numbers are involved and \Eref{eq:edge} can also be implemented by pixelwise value assignment operations, the extra parameter and computation overhead is extremely low and negligible. 

\begin{figure}[tp]
	\centering
	\includegraphics[width=122mm]{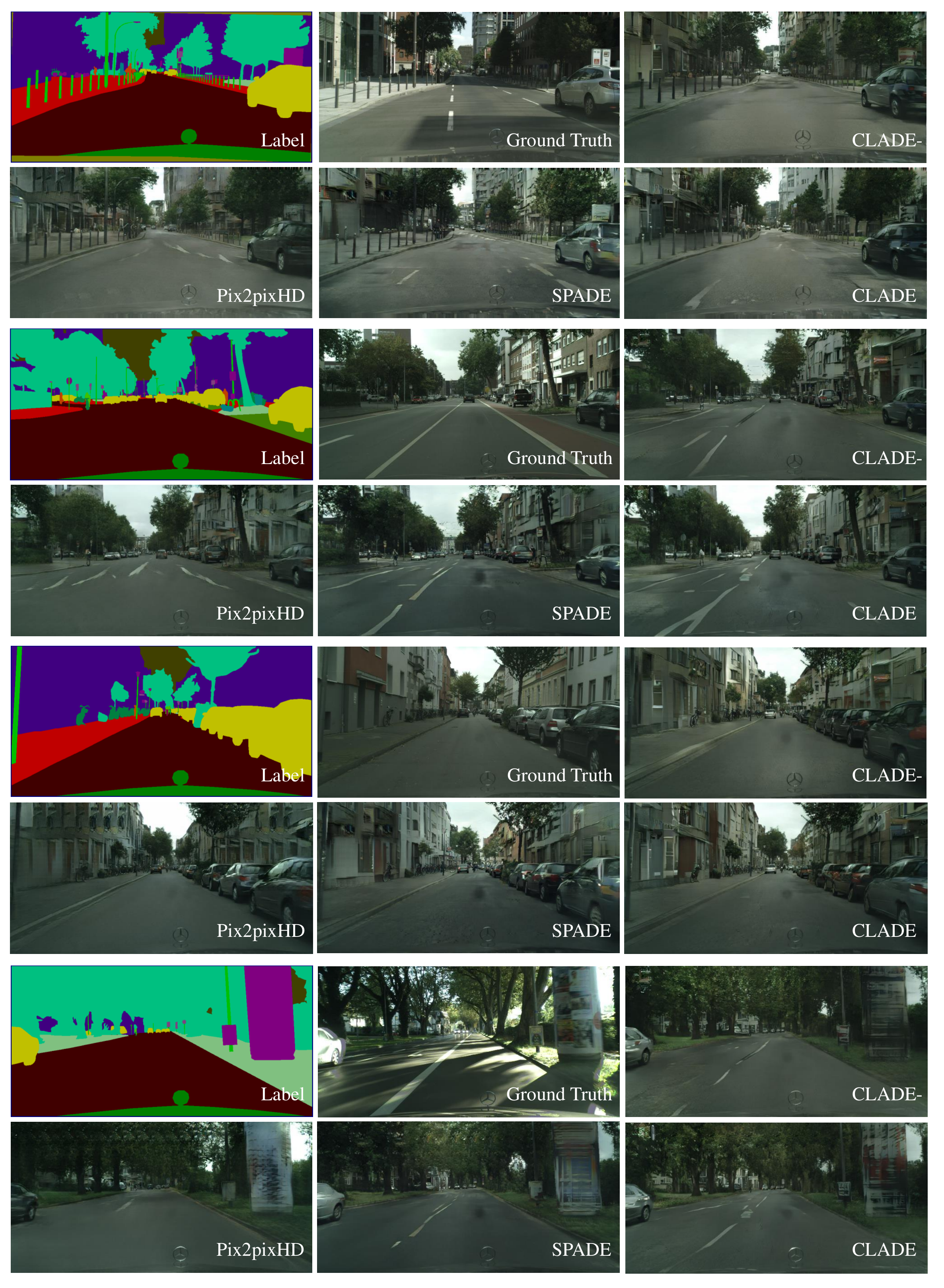}
	\caption{Visual comparison results on the Cityscapes dataset with the instance map as input. CLADE- represents the version that does not use the instance map.}
	\label{fig:sup_inst_city}
\end{figure}

\Tref{tab:metrics} reports the detailed experiment results on Cityscapes dataset. It can be seen that CLADE achieves comparable performance when compared to SPADE, and outperforms Pix2pixHD by a large margin. \Fref{fig:sup_inst_city} shows some qualitative comparison results. Our method works very well and separates close objects easily with the help of instance information. 
\begin{table}
	\centering
	\caption{Performance and complexity analysis on Cityscapes dataset with additional instance information input.}
	\begin{tabular}{c||c|c|c||c|c|c}
		\hline
		Method & mIoU & accu & FID & Params(M) & FLOPs(G) & Runtime(s) \\
		\hline
		Pix2pixHD & 51.73 & 91.59 & 69.98 & 182.53 & 605.29 & 0.128 \\
		SPADE & 67.50 & 94.82 & 54.36 & 93.05 & 1126.00 & 0.198 \\
		CLADE & 66.42 & 95.00 & 56.55 & 67.90 & 302.17 & 0.099 \\
	\end{tabular}
	\label{tab:metrics}
\end{table}

\section{More visual comparison results}
\begin{figure}[tp]
	\centering
	\includegraphics[width=122mm]{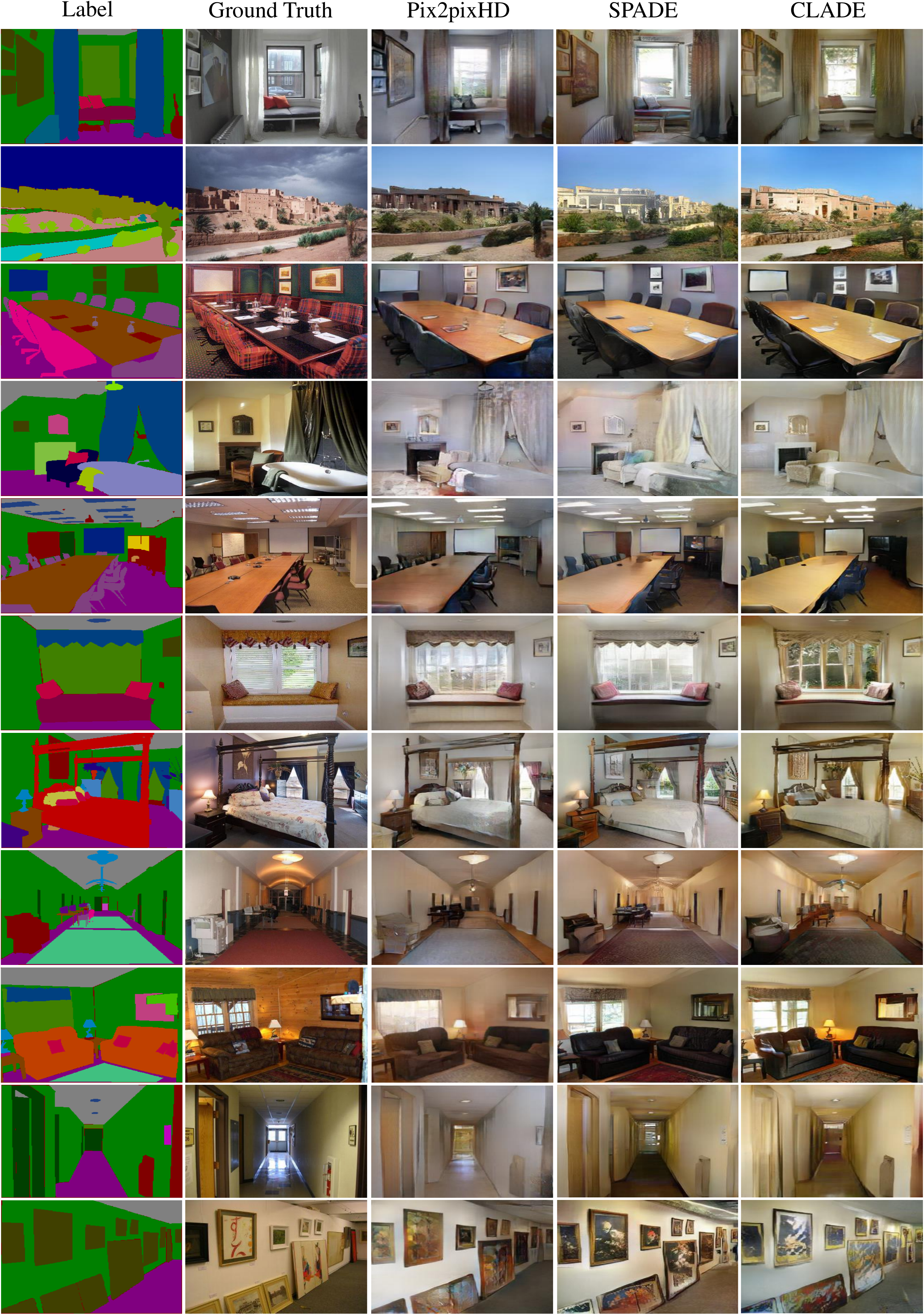}
	\caption{More visual comparison results on the ADE20k dataset. Our CLADE can generate high-quality images, which are comparable to SPADE and better than Pix2pixHD.}
	\label{fig:sup_ade}
\end{figure}

\begin{figure}[tp]
	\centering
	\includegraphics[width=122mm]{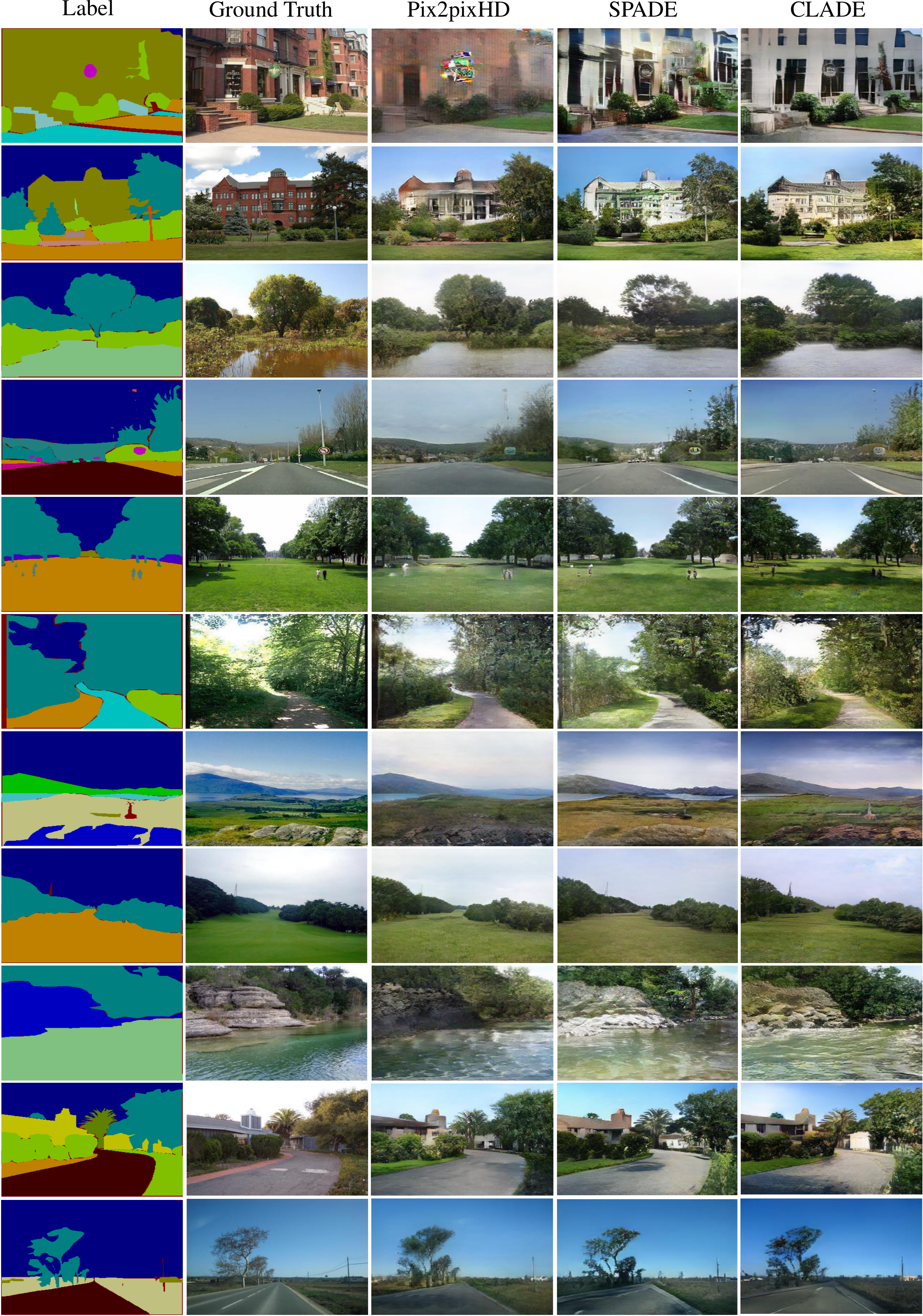}
	\caption{More visual results generated on the ADE20k-outdoor dataset, demonstrating the super performance of CLADE on large semantic regions with rich textures.}
	\label{fig:sup_ade_out}
\end{figure}

\begin{figure}[tp]
	\centering
	\includegraphics[width=122mm]{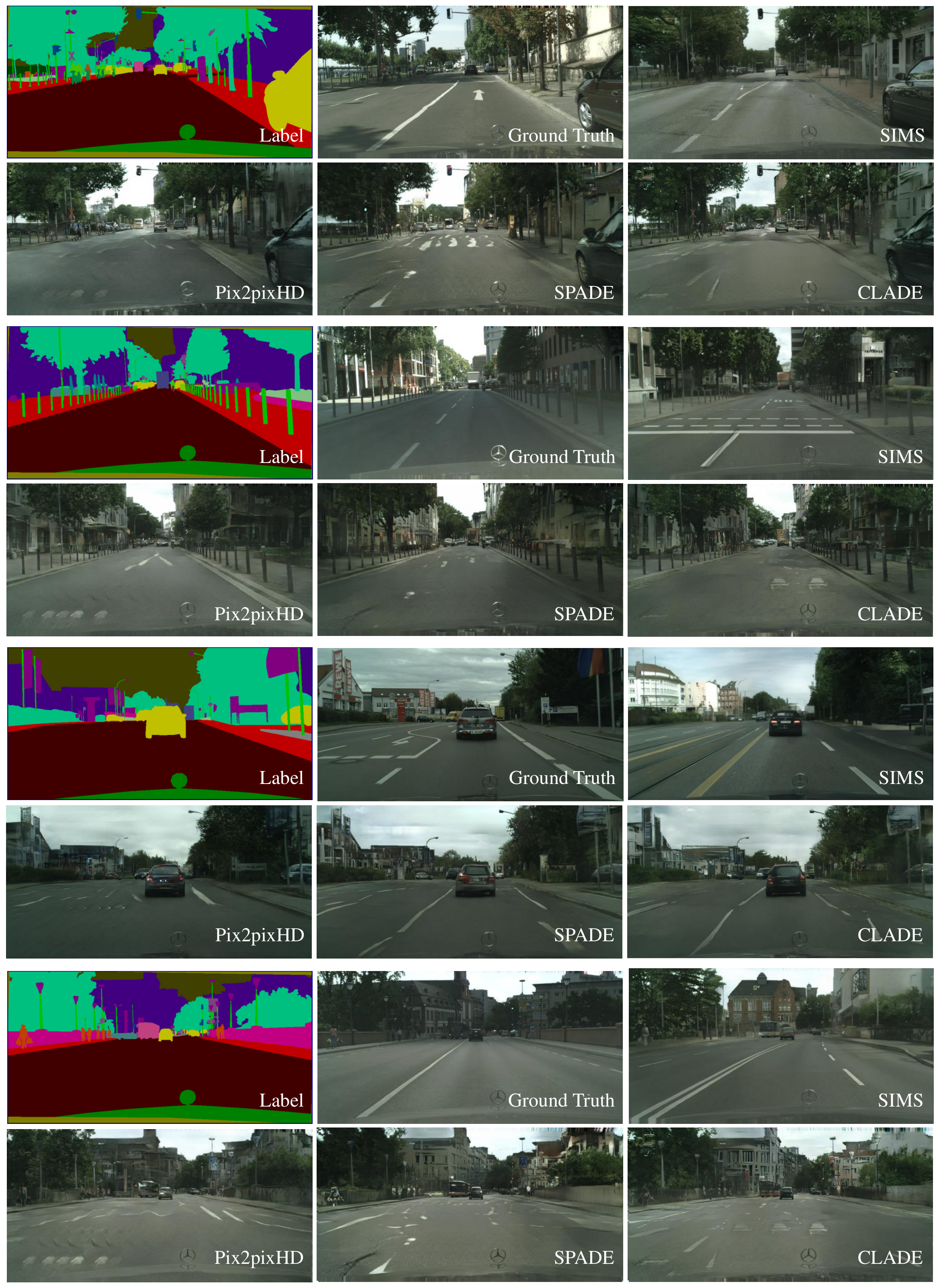}
	\caption{More visual results for high-dimensional image synthesis on the Cityscapes dataset. }
	\label{fig:sup_city}
\end{figure}

\begin{figure}[tp]
	\centering
	\includegraphics[width=122mm]{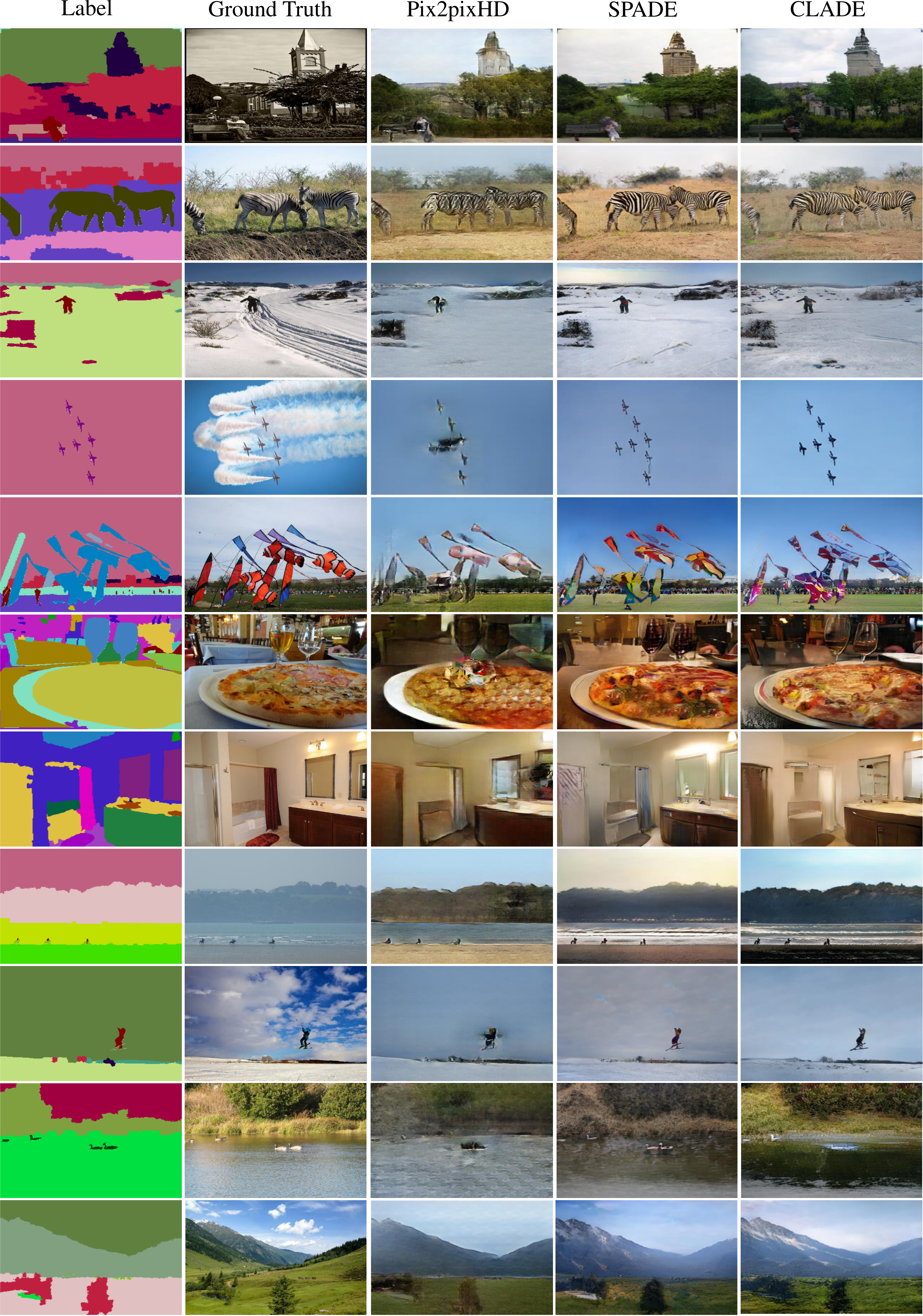}
	\caption{More visual results on the challenging COCO-Stuff dataset. It shows that our CLADE can produce visual pleasant results for very diverse categories.}
	\label{fig:sup_coco}
\end{figure}

In \Fref{fig:sup_ade},\ref{fig:sup_ade_out},\ref{fig:sup_city},\ref{fig:sup_coco}, more visual results are shown on the ADE20k, ADE20k-outdoor, Cityscapes and  COCO-stuff datasets respectively.

\end{document}